# CRACQ: A Multi-Dimensional Approach To Automated Document Assessment

Ishak Soltani [1], Francisco Belo,[1] Bernardo Tavares [1]

Granter.ai [1]

bernardo.tavares@granter.ai

### Abstract

This paper presents CRACQ, a multi-dimensional evaluation frame- work tailored to evaluate documents across five specific traits: Coherence, Rigor, Appropriateness, Completeness, and Quality. Building on insights from trait-based Automated Essay Scoring (AES), CRACQ expands its fo-cus beyond essays to encompass diverse forms of machine-generated text, providing a rubric-driven and interpretable methodology for automated evaluation. Unlike single-score approaches, CRACQ integrates linguistic, semantic, and structural signals into a cumulative assessment, enabling both holistic and trait-level analysis. Trained on 500 synthetic grant pro-posals, CRACQ was benchmarked against an LLM-as-a-judge and further tested on both strong and weak real applications. Preliminary results in-dicate that CRACQ produces more stable and interpretable trait-level judgments than direct LLM evaluation, though challenges in reliability and domain scope remain.

Keywords: Document quality assessment, LLM-as-a-judge, LoRA, CRACQ

## 1 Introduction

Evaluating open-ended document generation remains challenging in high-stakes settings such as grant proposal writing or long-form content creation. Existing paradigms include LLM-as-judge (LAAJ) pipelines (e.g. GPT-4 scoring) that produce human-like judgments but are opaque, often non-deterministic, and computationally [10] costly. Other approaches, such as judge-free benchmarks (e.g. pfgen-bench) and intrinsic metrics (e.g. GRUEN, MAUVE) [20, 15] apply distributional or n-gram-based methods to yield fast, deterministic reference- less scores. However, these methods yield only holistic quality scores without fine-grained, trait-specific diagnostics, limiting interpretability and actionable feedback. To address these gaps, we introduce CRACQ (Coherence, Rigor, Appropriateness, Completeness, Quality), a rubric-driven multi-trait evaluation framework that delivers scalable, interpretable trait-disaggregated assessment of generated documents like grant writing or long-form generation.

## 2 Related work

### 2.1 Automated Essay Scoring

Automated essay scoring has long emphasized holistic scoring, but recent work has shifted toward trait-specific evaluation [3]. State-of-the-art systems like TRATES and RMTS [2] leverage rubric-defined traits: TRATES uses a large language model (LLM) to convert each trait's rubric into targeted assessment questions, whose answers serve as interpretable features for that trait. Like- wise, RMTS prompts an LLM to generate trait-specific rationales based on rubric guidelines, which an essay scorer then uses to predict multi-trait scores [2]. These rubric-anchored methods yield detailed feedback aligned with human grading criteria, but they have been evaluated mainly on English essay cor-pora (e.g. ASAP++, ELLIPSE) their applicability to short-answer responses or non-English writing remains largely untested

### 2.2 General Document Evaluation, LLM Judges, and Benchmarking

Parallel lines of work use LLMs as general-purpose evaluation judges. For exam- ple, Themis (Hu et al., WWW 2025) [7] fine-tunes an LLM on human or high-end LLM ratings to produce context-aware evaluations that align closely with human preferences [8]. Other approaches (e.g. PandaLM, JudgeLM,) [17, 19] construct datasets from GPT-4 or human critiques to fine-tune open LLMs as scalable judge models. In contrast, judge-free benchmarks like pfgen-bench dispense with any LLM judge and instead use automatic metrics (e.g. n-gram-based fluency, truthfulness, helpfulness) that correlate with human/LLM ratings. Similarly, in- trinsic quality metrics such as GRUEN (evaluating grammaticality, coherence, focus, etc.) and MAUVE (measuring the statistical gap between machine and human text) provide reference-free assessments of generated text [20, 15], These LLM-based evaluation frameworks offer great scalability and nuance: they can process large volumes of outputs without human labor [12]. and small fine-tuned judges can nearly match very large models' judgments (e.g. Themis achieves
performance close to GPT-4 with $\ll 1\%$ of the parameters [7].However, they also have drawbacks. Evaluation results can be unstable or non-deterministic
(varying with model choice or prompt [5] and relying on opaque LLM APIs can hinder reproducibility and privacy. Moreover, LLM-based judgments incur high computational cost due to repeated model queries. CRACQ builds on these prior techniques by integrating rubric-informed trait scoring with LLM-based evaluation methods, aiming to deliver a multi-dimensional document evaluation framework that enhances interpretability and efficiency beyond existing approaches.

Figure 1: CRACQ Training and evaluation pipeline.

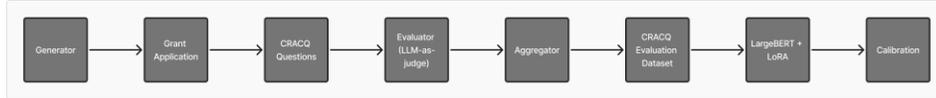

## 3 Methodology

### 3.1 Synthetic Data Generation

To obtain training data for grant evaluation, we constructed a synthetic corpus of applications seeded by 13 real grant prompts drawn from diverse funding domains. Using Granter.ai, an AI-driven grant-writing platform, we generated 500 unique proposals by conditioning on company background, project ideas, and specific grant guidelines. The prompts were carefully designed and validated by human grant experts to ensure realism and fidelity, resulting in a dataset that reflects authentic proposal structure and thematic variety while addressing the confidentiality barriers that limit access to real applications. In addition to this synthetic corpus, we also obtained four real grant applications, which are reserved for the evaluation phase to provide a complementary test of model performance under authentic conditions.

### 3.2 LAAJ Evaluation System for CRACQ dimensions

We use a LAAJ system as an automated judge to evaluate each synthetic proposal on multiple trait dimensions. While we acknowledge known limitations of LLM judges (e.g., variability with prompts, style biases), this choice is the most scalable way to obtain ne-grained scores without costly human annotation. Prior work [12] has shown that GPT-4 especially when prompted with chain-of-thought and rubric-like instructions can achieve good correlation with human judgments of text quality. Each application is evaluated on five CRACQ dimensions inspired by grant review criteria, following [9]

- Coherence: logical consistency and ow of the narrative.
- Rigor: thoroughness and soundness of arguments.
- Appropriateness: fit to the grant's goals and audience.

- Completeness: all required elements and objectives are addressed.
- Quality: overall writing quality and persuasiveness.

For each trait, we crafted specific questions so that LAAJ's answers would be rubric-anchored. In other words, we explicitly provided our LAAJ system with the criterion definition or probing questions for that trait, thus anchoring its judgment to the rubric [14]a. For example, under Coherence we asked,
Is the narrative logically organized with a consistent thread from beginning to end?;
under Completeness, Does the text clearly present the problem or opportunity it addresses, without leaving critical gaps? . We then applied a two-pass procedure for each document.

1. Structured evaluation: we feed the generated text back into a LAAJ sys- tem with a structured prompt containing the 25 trait-specific c questions (5 per CRACQ dimension). LAAJ returns a numeric score (0 1) and brief rationale for each question. By design, our system operates as both grader and explainer.

2. An aggregator for the scores: each dimension's score is the mean of its five question scores, and the overall CRACQ score is the mean of the five dimension scores.

## 3.3 Model Training

With LAAJ generated scores as training labels, we trained our automated eval-uator. We used the BigBird RoBERTa architecture [18], which employs sparse attention to handle long sequences (up to thousands of tokens). This model was implemented via Hugging Face's Transformers [18] the choice was due to it's long-form document encoding, BigBird-RoBERTa transformer capability to handle extended input sequences efficiently. BigBird's sparse attention mecha- nism scales linearly with sequence length, allowing it to process sequences up to 8 longer than the standard 512-token limit of typical Transformers. This en- ables the model to capture long-range dependencies in documents and has been shown to improve performance on tasks requiring extensive context. To adapt this pre-trained model to our task, we use Low-Rank Adaptation (LoRA) as a parameter-efficient ne-tuning strategy [6, 13] LoRA injects trainable low-rank update matrices into each Transformer layer while freezing the original weights, drastically reducing the number of trainable parameters needed for ne-tuning [6]. This approach retains the full model's performance despite updating only a small fraction of parameters, as evidenced by LoRA matching the quality of full ne-tuning with orders-of-magnitude fewer trainable parameters. The model outputs are produced via trait-specific regression heads, i.e., independent linear layers for each target trait [1]. This design allows each trait to be learned and calibrated separately, preserving trait-specific patterns and preventing interfer- ence among the predictions of different traits.

Table 1: CRACQ Model Performance Metrics

| Dimension | MAE | RMSE | Pearson Correlation |
|---|---|---|---|
| Coherence | 0.1064 | 0.1254 | 0.5583 |
| Rigor | 0.0928 | 0.1212 | 0.5443 |
| Appropriateness | 0.0911 | 0.1098 | 0.6335 |
| Completeness | 0.1083 | 0.1347 | 0.5420 |
| Quality | 0.1119 | 0.1302 | 0.5710 |
| Overall | 0.1021 | 0.1246 | 0.6807 |

### 3.3.1 Training Procedure and Calibration

Training used the AdamW [16] optimizer with a linear warm-up and decay schedule. We minimized a Huber loss [4] between predictions and the LAAJ scores, which is robust to outliers. We added a Pearson-correlation regular- ization term to preserve rank ordering of scores. After training, we applied an isotonic regression calibration [11] on a held-out validation set for each trait. This nonparametric calibration adjusts the model's output distribution to match the true score distribution, improving the reliability of final scores.

## 4 Results

### 4.1 CRACQ Model Performance Metrics

Table 1 represents a side-by-side comparison of LAAJ and CRACQ using RMSE, MAE, and Pearson correlation per trait and overall. At the trait level, RMSE for Coherence is ~3% higher than Rigor's RMSE (relative to Rigor's RMSE) Coherence shows slightly larger error magnitudes than Rigor yet both traits sit within 3% of the Overall RMSE (relative to the Overall RMSE), showing that they are representative of the system's average error. For MAE, Coherence is ~16% higher than Rigor/Appropriateness (relative to each baseline's MAE), i.e., its average absolute gap between systems is modestly larger. For corre- lation, differences are smaller: the Overall correlation is ~8% higher than the best trait-level correlation (relative to that trait's correlation), showing that the two systems move together more consistently when scores are averaged across dimensions than when considered trait by trait.

### 4.2 Comparison on 4 Real Grant Applications

Table 2 reports CRACQ's performance on the real grant corpus when compared against our LAAJ system. Across all five traits and the overall score, the model produced mean absolute errors in the range of 9 11%, while correlations with the reference judgments lay between 54% and 68%. Among the traits, Appro- priateness achieved the best alignment, with the lowest error and the highest
correlation (≈63%). Coherence and Quality also showed strong performance, both with errors close to 10% and correlations in the high-50% range. Rigor yielded a similarly low error but a somewhat weaker correlation, suggesting ac- curate absolute predictions yet less stable rank ordering. Completeness was the most di cult dimension, with error exceeding 13% and correlation in the lower- 50% range, reflecting the challenge of capturing full content coverage. At the
holistic level, CRACQ's overall score correlated strongly with the LAAJ system (≈68%) while keeping error near 10%

Table 2: Trait-Level Evaluation of a Real Grant

| Dimension | CRACQ | LAAJ |
|---|---|---|
| COHERENCE | 0.627 | 0.71 |
| RIGOR | 0.373 | 0.52 |
| APPROPRIATENESS | 0.533 | 0.62 |
| COMPLETENESS | 0.528 | 0.7 |
| QUALITY | 0.672 | 0.66 |
| OVERALL | 0.547 | 0.64 |

## 5 Conclusion

The empirical results also reflect the key contributions of this work. First, we created a new LLM-labeled dataset of document quality ratings, enabling repro- ducible evaluation research without depending on proprietary systems [5]. This addresses recent concerns about the reproducibility of LLMbased evaluations
[8] by providing a fixed reference set of judgments. Second, we implemented a long-context scoring model based on the BigBird Transformer and ne-tuned it with a low-rank adaptation strategy (LoRA). This efficient fine-tuning approach allowed CRACQ to handle full-length documents (thousands of tokens) while using limited computational resources. We trained the model with a robust Huber loss, which reduced the impact of outlier errors in the regression, and we applied an isotonic regression calibration to align the model's outputs with the true score distribution. These techniques improved the stability of trait predic- tions and ensured that the predicted scores remain interpretable as probabilities or ratings on the original scale. Third, we emphasize that CRACQ is an inter- pretable multitrait evaluator: instead of outputting a black-box score, it pro- vides a breakdown across coherent dimensions (Coherence, Rigor, etc.), which can help diagnose which aspects of a document need improvement. We have released the CRACQ model and evaluation framework publicly, so that other researchers can reproduce our results and build upon this multi-dimensional assessment approach.

### 5.1 Future Work and Outlook

The CRACQ framework demonstrates that automated models can approximate expert evaluators on complex, long-text quality assessment tasks. We view this as a foundational step toward more sophisticated automated document eval-uation. The current model's limitations point to clear directions for future work. Going forward, improvements could be achieved by incorporating more diverse training data (including human-labeled examples) to cover edge cases of each trait, and by exploring advanced architectures or adaptation methods to enhance trait-specific sensitivity. For example, integrating domain-specific c knowledge or iterative self-re nement might boost the model's ability to judge completeness and rigor. Additionally, future research should investigate tech- niques to further narrow the gap between CRACQ's predictions and human or LLM judges potentially by calibrating against human scores directly or by using ensemble approaches to improve consistency.

Table 3: Trait-Level Evaluation of a Real Grant

| Dimension | CRACQ | LAAJ |
|---|---|---|
| COHERENCE | 0.612 | 0.71 |
| RIGOR | 0.481 | 0.58 |
| APPROPRIATENESS | 0.544 | 0.68 |
| COMPLETENESS | 0.60 | 0.75 |
| QUALITY | 0.72 | 0.74 |
| OVERALL | 0.591 | 0.69 |

Table 4: Trait-Level Evaluation of a Real Grant

| Dimension | CRACQ | LAAJ |
|---|---|---|
| COHERENCE | 0.658 | 0.74 |
| RIGOR | 0.542 | 0.46 |
| APPROPRIATENESS | 0.521 | 0.59 |
| COMPLETENESS | 0.603 | 0.66 |
| QUALITY | 0.69 | 0.64 |
| OVERALL | 0.603 | 0.61 |

# 6 Appendix

```
"quality":
        "is the writing professional, polished, and free from errors that harm readability?",
        "is the problem framed with clear context and relevance?",
        "is the structure organized to support readability and persuasiveness?",
        "does the text show originality and depth through precise, distinctive expression?",
        "is the overall presentation fluent, persuasive, and convincing in language and
structure?"

"completeness":
        "Does the text clearly present the problem or opportunity it
addresses, without leaving critical gaps?",
        "Are the goals and success criteria described explicitly and
traceably throughout the text?",
        "Is the proposed solution described with enough linguistic and conceptual detail to be
understandable and reproducible?",
        "Does the document include a structured plan (milestones or sequencing)
that is easy to follow?",
        "Are dependencies, resources, and constraints explicitly
```

Table 5: Trait-Level Evaluation of a Real Grant

| Dimension | CRACQ | LAAJ |
|---|---|---|
| COHERENCE | 0.711 | 0.68 |
| RIGOR | 0.643 | 0.72 |
| APPROPRIATENESS | 0.673 | 0.78 |
| COMPLETENESS | 0.582 | 0.73 |
| QUALITY | 0.668 | 0.68 |
| OVERALL | 0.655 | 0.71 |

```
acknowledged and integrated into the description?"

"coherence":
    "Is the narrative logically organized with a consistent thread from beginning to end?",
    "Do the objectives, methods, and outcomes align without logical breaks?",
    "Are assumptions stated explicitly and referenced consistently,
avoiding contradictions?",
    "Is the writing clear, with sentences and paragraphs connected cohesively?",
    "Is the overall argument persuasive and internally consistent in tone and reasoning?"

"rigor":
    "Is the methodology described with clarity, precision, and justification?",
    "Does the text acknowledge risks, biases, or uncertainties in realistic language?",
    "Are metrics or evaluation methods expressed precisely and in a replicable way?",
    "Is prior work or evidence referenced to strengthen the argument?",
    "Are limitations acknowledged with a credible discussion of how they are handled?"

"appropriateness":
    "Is the tone, vocabulary, and style appropriate for a professional proposal?",
    "Is the scope linguistically realistic given the described
resources and timeframe?",
    "Is the intended audience addressed with a suitable register
(neither too casual nor too opaque)?",
    "Are ethical, safety, or compliance concerns handled with
responsible language?",
    "Is ambition framed with language that balances aspiration and feasibility?"
```